\documentclass{article}

\usepackage{PRIMEarxiv}

\usepackage[utf8]{inputenc} % allow utf-8 input
\usepackage[T1]{fontenc}    % use 8-bit T1 fonts
\usepackage{url}            % simple URL typesetting
\usepackage{booktabs}       % professional-quality tables
\usepackage{amsfonts}       % blackboard math symbols
\usepackage{nicefrac}       % compact symbols for 1/2, etc.
\usepackage{microtype}      % microtypography
\usepackage{lipsum}
\usepackage{fancyhdr}       % header
\usepackage{graphicx}       % graphics
\graphicspath{{media/}}     % organize your images and other figures under media/ folder

\usepackage{natbib}
\usepackage{wrapfig}
\usepackage{enumitem}
\usepackage{amsmath}
\usepackage{multirow}
\usepackage{booktabs}
\usepackage{caption}
\usepackage{algorithm}
\usepackage{algpseudocode}
\usepackage[dvipsnames,table]{xcolor}
\definecolor{darkgreen}{RGB}{0,128,0}
\usepackage[breakable]{tcolorbox}
\usepackage{geometry}
\usepackage{hyperref}       % hyperlinks
\title{Learning from Failures: Correction-Oriented Policy Optimization with Verifiable Rewards}

\author{
\textbf{Mengjie Ren}\textsuperscript{1,2}, 
\textbf{Jie Lou}\textsuperscript{3}, 
\textbf{Boxi Cao}\textsuperscript{1}, 
\textbf{Xueru Wen}\textsuperscript{1,2}, 
\textbf{Hongyu Lin}\textsuperscript{1}, 
\textbf{Xianpei Han}\textsuperscript{1}, 
\textbf{Le Sun}\textsuperscript{1}, 
\textbf{Xing Yu}\textsuperscript{3}, 
\textbf{Yaojie Lu}\textsuperscript{1} \\
\\
\textsuperscript{1}Chinese Information Processing Laboratory, Institute of Software, Chinese Academy of Sciences \\
\textsuperscript{2}University of Chinese Academy of Sciences \\
\textsuperscript{3}Xiaohongshu Inc.
}

\date{} 

\begin{document}
\maketitle
\begin{abstract}
Reinforcement Learning with Verifiable Rewards (RLVR) has emerged as an effective paradigm for improving the reasoning capabilities of large language models.
However, RLVR training is often hindered by sparse binary rewards and weak credit assignment, resulting in ambiguous optimization signals and underutilization of the useful information embedded in failed trajectories.
To address this challenge, we propose \textit{\textbf{C}orrection-Or\textbf{I}ented \textbf{P}olicy \textbf{O}ptimization} (CIPO), a simple and effective extension to RLVR that converts on-policy failed trajectories into correction-oriented supervision, without relying on any external signals.
By jointly optimizing correction samples derived from the model’s own failed attempts together with the standard RLVR objective, CIPO improves learning effectiveness while explicitly enhancing the model’s ability to correct its own errors.
Extensive experiments across 11 benchmarks spanning mathematical reasoning and code generation demonstrate that CIPO consistently and significantly outperforms strong baselines in both reasoning and correction performance.
Moreover, CIPO yields stronger pass@K gains, indicating that it improves the model’s intrinsic reasoning capacity rather than merely redistributing probability mass over existing correct answers.
\end{abstract}

% keywords can be removed
% \keywords{First keyword \and Second keyword \and More}

\section{Introduction}
Reinforcement Learning with Verifiable Rewards (RLVR) has emerged as a core paradigm for enhancing the reasoning capabilities of large language models (LLMs), with notable success in mathematical reasoning and code generation~\citep{openai2024openaio1card,Guo_2025,kimiteam2025kimik15scalingreinforcement}. 
By leveraging automatically verifiable reward signals from on-policy rollouts, RLVR enables scalable training without requiring additional human annotations.

\begin{wrapfigure}{r}{0.5\textwidth}  
  \centering
  \includegraphics[width=\linewidth]{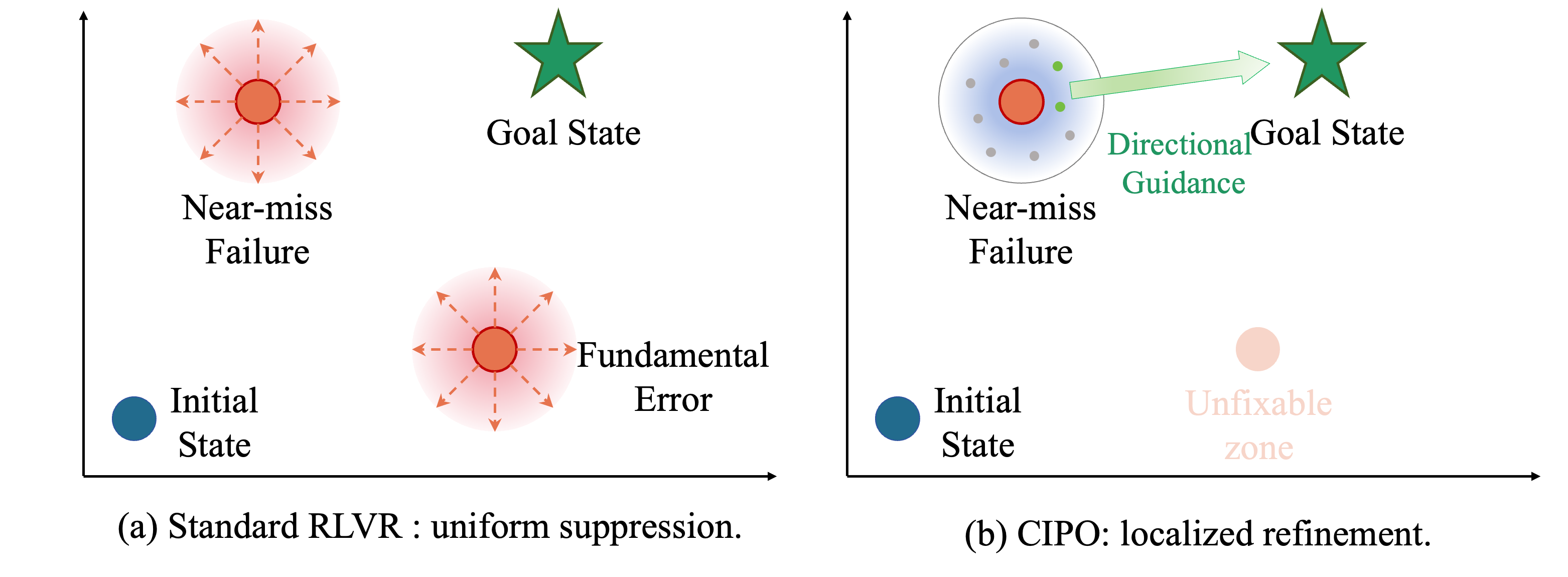}
  \caption{Comparison of how standard RLVR and CIPO exploit failed trajectories. CIPO provides more directional and informative learning signals.}
  \label{fig:example}
\end{wrapfigure}
 % by converting failures into correction-oriented supervision
 
Despite the success, existing RLVR algorithms such as Group Relative Policy Optimization (GRPO)~\citep{shao2024deepseekmathpushinglimitsmathematical} are fundamentally built upon a reinforce–suppress paradigm, where successful trajectories are reinforced while failed ones are uniformly penalized, regardless of their logical proximity to the ground truth~\citep{hubotter2026reinforcement}.
Due to the binary and sparse nature of verifiable rewards, training signals often provide ambiguous optimization guidance and fail to capture the heterogeneous nature of failures, particularly in long-horizon reasoning. As illustrated in Figure~\ref{fig:example}(a), failed rollouts may arise from fundamentally different error modes, ranging from critical logical flaws and intermediate inconsistencies to minor final-step miscalculations.
By treating all failures as identical negative signals, existing approaches merely suppress the likelihood of entire trajectories, without offering explicit guidance on how specific errors can be corrected~\citep{yue2025doesreinforcementlearningreally}. 
Moreover, failed trajectories often contain partially correct reasoning steps that constitute valuable learning signals. Discarding such intermediate structures not only wastes useful supervision but may also hinder effective exploration, ultimately leading to suboptimal generalization~\citep{hu2026rewarding,yue2025doesreinforcementlearningreally,hubotter2026reinforcement}.

Previous studies have sought to address these challenges through the integration of additional process reward models~\citep{cui2025process,wang2024math} or LLM-based critics~\citep{xie2025capo}. Nevertheless, these methods are often hampered by the costs of additional manual labeling and computation resources, while the limited capacity of the auxiliary models can introduce noise and undermine generalizability~\citep{wen2024rethinking,gao2023scaling}. 
More recently, such as SDPO~\citep{hubotter2026reinforcement}, leverages environmental feedback or self-generated trajectories to construct a feedback-conditioned teacher and derive fine-grained supervision from distributional discrepancies. However, these methods rely on reliable feedback signals and reflective capabilities that are often limited in weaker models.
Moreover, its generalization has been criticized for suppressing epistemic uncertainty, thereby undermining robust reasoning~\citep{kim2026does}.
Consequently, there is an urgent need for a task-agnostic solution that addresses these challenges without requiring additional external supervision signals.

To this end, we propose \textit{\textbf{C}orrection-Or\textbf{I}ented \textbf{P}olicy \textbf{O}ptimization}~(\textsc{CIPO}), a systematic extension within the RLVR paradigm without requiring any external information. 
The core idea of CIPO is to transform on-policy failed trajectories from mere objects of penalty into exploitable supervisory signals. 
Specifically, in figure~\ref{fig:framework}, during each policy update, we construct correction pairs from failed trajectories by conditioning the model on the original prompt together with its own erroneous output, and then sampling refined solutions. This correction objective is then jointly optimized with the standard GRPO objective.
Since all correction samples are derived from the model’s own on-policy failures without additional human annotation, CIPO ensures strict consistency between the training and inference distributions.
Furthermore, to prevent policy degradation caused by naively incorporating all failed trajectories into training, we integrate an adaptive mechanism that dynamically balances the proportion of successful versus failed trajectories, along with risk-aversion reward shaping. 
Moreover, we design a rollout preference strategy based on on-policy sampling accuracy to ensure a sustained and informative training signal.
These designs enable CIPO to effectively exploit the information contained in failed samples while preserving the original advantages of RLVR.

Intuitively, as illustrated in Figure~\ref{fig:example},\textsc{CIPO} improves RLVR from two complementary perspectives. First, the correction objective provides learning signals with stronger directionality. 
Crucially, this process differentiates failure modes by sampling in the local neighborhood of erroneous trajectories: a ``near-miss'' attempt (e.g., simple final-step calculation errors) has a much higher probability of yielding correct solutions during refinement sampling than a fundamentally flawed one. 
By naturally leveraging these varying rectification probabilities, \textsc{CIPO} extracts richer, denser signals from failures, reducing gradient ambiguity.
Second, \textsc{CIPO} explicitly trains the model's correction capability, generating correct solutions conditioned on its own erroneous attempts.
This enables our trained model not only to improve its reasoning ability but also to acquire stronger error-correction skills, thereby extending its practical applicability to scenarios such as debugging and refinement.

We conduct extensive experiments across 11 representative benchmarks spanning mathematical reasoning and code generation. Results show that \textsc{CIPO} consistently improves both reasoning and error-correction performance over strong baselines.
For correction, Seed-Coder-8B~\citep{seed2025seedcoderletcodemodel} trained with \textsc{CIPO} achieves a 7.63\% gain on DebugBench~\citep{tian-etal-2024-debugbench}, reaching performance comparable to Claude-4-sonnet~\citep{anthropic_claude4_news} and surpassing GRPO.
For reasoning, Qwen-3-4B~\citep{yang2025qwen3technicalreport} trained with \textsc{CIPO} improves average accuracy by 17.56\% across six mathematical benchmarks, outperforming GRPO by 4.55\%.
Additionally, \textsc{CIPO} yields higher pass@K, suggesting that it goes beyond simple probability concentration, thereby enhancing intrinsic reasoning~\citep{yue2025doesreinforcementlearningreally}.
In summary, our contributions are:

\begin{itemize}[leftmargin=*, itemsep=0pt]
    \item We revisit the role of failed trajectories in RLVR and investigate how they can be transformed from sparse negative feedback into useful correction-oriented supervision.
    \item We propose \textsc{CIPO}, a correction-oriented extension for RLVR that constructs correction samples from on-policy failed trajectories without additional annotations.
    \item Extensive experiments across 11 benchmarks demonstrate that \textsc{CIPO} consistently outperforms strong baselines in both reasoning and correction tasks, with further gains in pass@K metrics indicating genuine expansion of reasoning capabilities rather than probability redistribution.
\end{itemize}

\section{Preliminaries}

% In this section, we briefly introduce the reinforcement learning with verifiable rewards (RLVR) paradigm and review Group Relative Policy Optimization (GRPO), focusing on how successful and failed trajectories are utilized.
In this section, we briefly introduce RLVR and review GRPO, a representative algorithm in this paradigm.

\subsection{Reinforcement Learning with Verifiable Rewards}

RLVR is a paradigm tailored for LLM reasoning tasks where the validity of generated outputs can be automatically verified—for instance, checking the final answer in mathematical reasoning or functional execution in code generation.

Given a prompt $x \sim \mathcal{D}$, a policy $\pi_\theta$ generates a rollout $y$ autoregressively and receives a binary reward $R(x,y) \in \{0,1\}$. The objective of RLVR is to maximize the expected reward:
\[
\max_\theta \; \mathbb{E}_{x \sim \mathcal{D},\, y \sim \pi_\theta(\cdot|x)} \big[ R(x,y) \big].
\]
Due to the sparse and sequence-level nature of verifiable rewards, policy optimization in RLVR typically relies on sampling-based gradient estimators.

\subsection{Group Relative Policy Optimization}

GRPO is designed to stabilize training under sparse binary rewards without requiring a value model. For each prompt $x$, GRPO samples a group of $N$ trajectories $\{y_i\}_{i=1}^N$ from the current policy and evaluates their rewards $\{r_i\}_{i=1}^N$.
GRPO computes a normalized relative advantage within each group:
\[
A_i = \frac{r_i - \mu_r}{\sigma_r}, 
\quad
\mu_r = \frac{1}{N} \sum_{j=1}^N r_j,
\]
where $\sigma_r$ denotes the standard deviation of rewards in the group. The policy is updated by reinforcing trajectories with positive advantages and suppressing those with negative advantages.

Under this formulation, successful trajectories are reinforced relative to the group mean. However, failed trajectories receive uniformly negative advantages whenever successful trajectories exist in the group, regardless of their specific error modes or potential partial correctness.

\section{Correction-Oriented Policy Optimization}
\begin{figure*}[!t]
    \centering
    \includegraphics[width=0.90\textwidth]{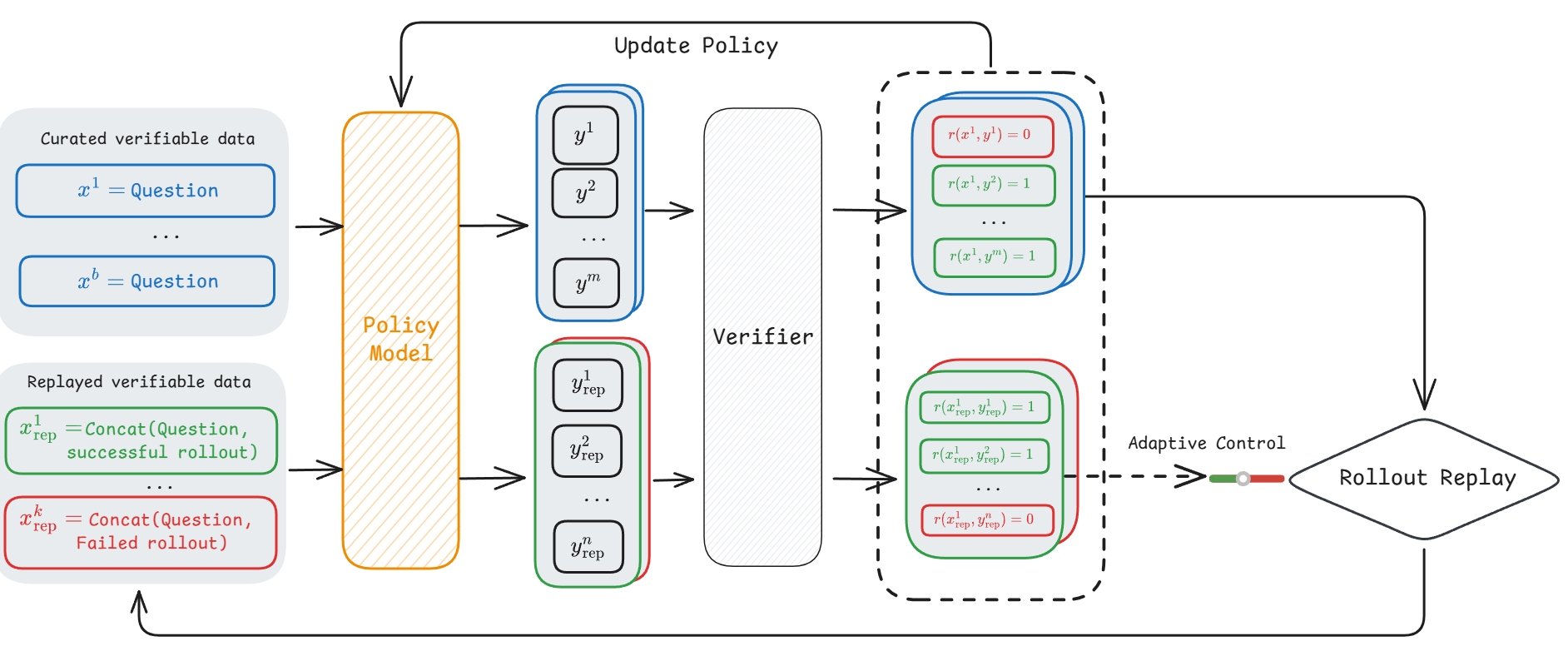}
    \caption{The overall framework of \textsc{CIPO}. First, we generate rollouts for the curated data via the policy model and verify their correctness. Subsequently, we construct replayed samples using a template governed by an adaptive mechanism, which dynamically adjusts the ratio of successful to failed rollouts in the replay. We then generate and verify rollouts for this replayed data. Finally, we perform RL on the rollouts from both the replayed and original samples.}
    \label{fig:framework}
    % \vskip -2pt
\end{figure*}

To address the aforementioned limitations of current RLVR methods, we propose \textsc{CIPO}, which transforms on-policy failed trajectories from mere objects of penalty into exploitable supervisory signals. In this section, we first introduce the overall procedure of \textsc{CIPO} (§\ref{sec:training_procedure}), then describe two key strategies designed to enhance training stability and efficiency: adaptive replay with risk-averse shaping (§\ref{sec:adaptive_replay}) and difficulty-aware trajectory preference (§\ref{sec:medium_hard}). The core algorithm is outlined in Appendix~\ref{appendix:Details-about-method}.

\subsection{Overall Procedure}
\label{sec:training_procedure}

The overall framework of \textsc{CIPO}, illustrated in Figure~\ref{fig:framework}, extends standard RLVR by establishing an iterative cycle of generation and \textit{correction-oriented replay}. At each training step $t$, we optimize the policy $\pi_\theta$ using two data streams: (1) \textbf{Base Stream}: Standard on-policy rollouts ${y_i}$ generated from original queries $x \sim \mathcal{D}$; (2) \textbf{Correction Stream}: Refinement rollouts ${y'_i}$ generated by conditioning the policy on the original query and a previous trajectory $y$ (i.e., prompts $x_{\text{rep}} = \text{Concat}(x, y)$; the concatenation template is detailed in Appendix~\ref{sec:Correction Prompt Construction}).

\textbf{From Suppression to Directional Guidance.} Standard RLVR methods (e.g., GRPO) inefficiently treat all failures with uniform negative suppression, providing no information on \textit{how} to improve. \textsc{CIPO} transforms these failures into informative anchors. By successfully refining a specific error $y_{fail}$ into a correct solution $y'$, the model establishes a distinct gradient path connecting the failure mode to the goal state as shown in Figure~\ref{fig:example}(b). This converts ambiguous suppression signals into precise directional guidance. However, indiscriminately training on all failed trajectories introduces severe distribution shift and learning inefficiencies. To mitigate these risks, we introduce two main strategic mechanisms.

\subsection{Adaptive Replay with Risk-Averse Shaping}

\label{sec:adaptive_replay}

To prevent policy degradation caused by naively incorporating all failed trajectories into training, we propose two complementary mechanisms for stable and efficient learning: \emph{adaptive replay ratio}, which dynamically adjusts the mixture of successful and failed trajectories, and \emph{risk-averse reward shaping}, which explicitly penalizes capability regressions.

\textbf{Adaptive Replay Ratio.} To balance learning from failed trajectories with retaining previously acquired capabilities, we maintain a dynamic replay ratio $\rho_t \in [\rho_{\min}, \rho_{\max}]$ for mixing successful and failed trajectories. This ratio is adjusted according to the model's recent retention performance on recycled successful samples: when performance degrades or continues to decline, we increase the replay fraction of successful trajectories; when performance remains stable and high, we allow more emphasis on failed trajectories. This yields a simple feedback-based replay mechanism, with the full update rule deferred to Appendix~\ref{appendix:updateratio}.

\textbf{Risk-Averse Reward Shaping.} Inspired by risk-sensitive reinforcement learning~\citep{mihatsch2002risk}, we introduce an asymmetric penalty mechanism to impose a stronger constraint against capability regressions. Although adaptive mixing can adjust the correctness distribution of replayed rollouts, it does not directly penalize the following failure mode: the model is conditioned on a correct trajectory yet generates an incorrect response. To mitigate this issue, we impose an additional penalty on ``correct $\rightarrow$ incorrect'' transitions:
\begin{equation}
R_{\text{risk}}(x, y, y') = R(x, y') 
- \lambda_{\text{risk}} \cdot \mathbb{I}[R(x, y)=1 \land R(x, y')=0]
\label{eq:risk_shaping}
\end{equation}
where $y$ denotes the conditioning trajectory and $y'$ is the new response. This penalty is activated when the conditioning trajectory is correct but the new response is incorrect. In this way, the objective explicitly suppresses capability regressions, prioritizing the preservation of existing correct behaviors while still enabling the acquisition of new ones.

The combination of adaptive replay and risk-averse reward shaping creates a self-regulating training system: the adaptive controller manages the curriculum at a macro level by adjusting trajectory composition, while the shaped reward provides micro-level guidance by penalizing individual regressions. Together, these mechanisms enable stable learning from failure while preserving the model's ability to reproduce correct solutions when conditioned on them.

\subsection{Difficulty-Aware Trajectories Preference}
\label{sec:medium_hard}
% 1. Core Idea: To [Goal], we propose [Method], thereby [Result].
To improve learning efficiency, we propose a Difficulty-aware Trajectories Preference mechanism that prioritizes replaying prompts with moderate pass rates, thereby ensuring the model focuses on the effective learning window.
% 2. Explanation (Citations + User's specific phrasing)
Previous studies~\citep{yu2025dapoopensourcellmreinforcement, cui2025process, li2025questaexpandingreasoningcapacity,chen2025self} indicate that prompts that are consistently solved (too easy) or consistently failed (too hard) may hinder the learning process or contribute zero gradient signals. Replaying such samples wastes computational resources.

% 3. Specific Implementation
Specifically, we target the medium-difficulty regime. We define the set of prioritized prompts $\mathcal{X}_{\text{med}}$ as:
\begin{equation}
\mathcal{X}_{\text{med}} = \{ x \in \mathcal{D} \mid \delta_{\text{low}} \le \hat{P}(x) \le \delta_{\text{high}} \}
\end{equation}
where $\hat{P}(x)$ represents the empirical pass rate, and $\delta_{\text{low}}, \delta_{\text{high}}$ are thresholds. When insufficient medium-difficulty prompts are available, we adopt a fallback strategy that samples from the full distribution $\mathcal{D}$ (see Algorithm~\ref{alg:replay} in Appendix~\ref{appendix:Details-about-method}).

\subsection{Training Objective}
\label{sec:objective}

The joint objective combines base and correction rollouts:
\begin{align}
\mathcal{J}_{\text{CIPO}}(\theta)
&= \mathbb{E}_{x \sim \mathcal{D}} \Bigg[
\frac{1}{m} \sum_{i=1}^m A^{(i)} \log \pi_\theta\!\big(y^{(i)} \mid x\big)
\Bigg]
% \nonumber \\
+ \lambda \mathbb{E}_{(x,y_c,r_c) \sim \mathcal{X}_{\text{rec}}} \Bigg[
\frac{1}{n} \sum_{i=1}^n A'^{(i)} \log \pi_\theta\!\big(y'^{(i)} \mid x, y_c\big)
\Bigg]
\end{align}
where advantages are computed separately within each group, and correction rewards incorporate risk-averse shaping. $m$ and $n$ denote the numbers of sampled responses for base and correction rollouts, respectively, while $\lambda > 0$ controls the relative importance of correction rollouts. The core algorithm is summarized in Algorithm~\ref{alg:cipo}.

\section{Experiments}
\label{sec:experiments}

% We demonstrate the effectiveness of our approach on two verifiable domains: mathematical reasoning and code generation.

\subsection{Setup}

\textbf{Training Dataset}
For mathematical reasoning, following previous works~\citep{li2025jointlyreinforcingdiversityquality}, we utilize the DeepScalerR~\citep{anonymous2025deepscaler}, which consists of approximately 40,000 unique mathematics problem-answer pairs. For code generation, we curate verifiable prompts from AM-DeepSeek-Distilled-40M~\citep{tian2025deepdistillenhancingllmreasoning} with a primary focus on Python code generation and obtain approximately 370,000 unique items that can be verified by our sandbox server~\citep{bytedanceseedfoundationcodeteam2025fullstackbenchevaluatingllms}.

\textbf{Baselines and Variants}
\textsc{CIPO} is orthogonal to existing open-source RL training recipes and can be integrated with various base algorithms. 
In this work, we instantiate \textsc{CIPO} on top of GRPO and compare against vanilla GRPO under different training budgets as the baseline. We also compare with PRIME~\citep{cui2025process} under RLOO~\citep{ahmadian2024back}, which adheres to its official implementation.
% and SDPO~\citep{hubotter2026reinforcement}, which use tokenized feedback or successful rollout during training for logit-level credit assignment.
Additionally, to isolate the contribution of online replay, we report an offline variant that only replays trajectories collected at initialization rather than continuously during training.

\textbf{Implementation} We use the instruct mode of Qwen3-4B~\citep{yang2025qwen3technicalreport} for math experiments and Seed-Coder-8B~\citep{seed2025seedcoderletcodemodel} for code experiments. 
We implement our RL training pipeline with the \textsc{verl} framework~\citep{sheng2024hybridflow}. Each batch contains 128 questions, and we generate 8 responses per question during rollout. For rollout sampling, we use temperature $=1.0$, top-$p$ $=1.0$, and a maximum of 4096 tokens. We set the learning rate to $1\times 10^{-6}$ and KL loss coefficient to $1 \times 10^{-4}$. All models are trained for 500 steps, and we report results on the final step~\footnote{PRIME exhibits early training instability and fail to maintain stable optimization up to 500 steps.  We therefore report their best-performing checkpoints for a fair comparison.}. For \textsc{CIPO}, we set $\lambda=1$, the correction batch size to 128 and the number of correction rollouts to 8.

\textbf{Benchmarks}
We evaluate our method on diverse reasoning benchmarks with a maximum generation length of 8192 tokens.
\textit{Math.} We evaluate on AIME24/25~\citep{aime24,aime25}, AMC23, MATH500~\citep{lightman2023lets}, Minerva~\citep{lewkowycz2022solving}, and OlympiadBench~\citep{he-etal-2024-olympiadbench}.
For math datasets with fewer than 100 problems, we use temperature sampling (temperature=0.7) with 32 samples per problem and report pass@1.
For larger datasets, we use greedy decoding.
\textit{Coding.}
We evaluate on LiveCodeBench v6~(2024.8--2025.5)~\citep{jain2024livecodebench}, and LeetCode problems collected by DebugBench~\citep{tian-etal-2024-debugbench}, with unit tests from~\citep{xia2025leetcodedataset} due to the unavailability of automated official submission.
Following the official setting, we run LiveCodeBench 10 times with temperature=0.2, and use greedy decoding for LeetCode.
\textit{Correction.}
We evaluate on CriticBench~\citep{lin-etal-2024-criticbench} under three completeness settings using greedy decoding.
For DebugBench~\citep{tian-etal-2024-debugbench}, we use temperature=0.2 and run 8 times.

\begin{table*}[!t]
\centering
\caption{Main results on mathematical reasoning and code generation benchmarks. Mathematical reasoning is evaluated on Qwen3-4B, and code generation on Seed-Coder-8B.}
\label{tab:main_results}
\small
\setlength{\tabcolsep}{1pt}
\begin{tabular}{@{}lcccccccccc@{}}
\toprule
& \multicolumn{7}{c}{{Math}}
& \multicolumn{3}{c}{{Code}} \\
\cmidrule(lr){2-8} \cmidrule(lr){9-11}
Method
& AIME24 & AIME25 & AMC23 & MATH500 & Minerva & Olympiad & \multirow{2}{*}{\textbf{Avg}}
& LCBv6 & LeetCode & \multirow{2}{*}{\textbf{Avg}} \\
& Avg@32 & Avg@32 & Avg@32 & Avg@1 & Avg@1 & Avg@1 &
& Avg@10 & Avg@1 & \\
\midrule
Initial
& 23.54 & 21.04 & 66.02 & 82.20 & 40.44 & 47.70 & 46.82
& 24.12 & 69.16 & 46.64 \\
\addlinespace[0.5ex]
PRIME
& \textbf{48.65} & 40.42 & 85.31 & 91.20 & \textbf{54.04} & 57.63 & 62.88
& 26.65 & 69.43 & 48.04 \\
\addlinespace[0.5ex]
GRPO$_{\tiny \text{BS}=128}$
& 42.08 & 35.73 & 85.55 & 90.02 & 47.79 & 57.63 & 59.83
& 29.45 & 76.23 & 52.84 \\
GRPO$_{\tiny \text{BS}=256}$
& 40.42 & 33.44 & 84.06 & 90.00 & 51.10 & 58.96 & 59.66
& 29.32 & 75.93 & 52.63 \\
\addlinespace[0.5ex]
GRPO$^{\scriptsize m=16}_{\scriptsize \mathrm{BS}=128}$
& 44.38 & 37.40 & 85.70 & 89.00 & 53.68 & 56.44 & 61.10
& 29.01 & 71.08 & 50.05 \\
\rowcolor{cyan!10}
\addlinespace[0.5ex]
{CIPO}
& 47.50 & \textbf{44.90} & \textbf{89.61} & \textbf{92.00}
& 52.57 & \textbf{59.70} & \textbf{64.38}
& \textbf{30.33} & \textbf{78.21} & \textbf{54.27} \\
\bottomrule
\end{tabular}
\end{table*}

\begin{table}[t]
\begin{minipage}[t]{0.53\linewidth}
    \vspace{0pt}
    \centering
    \captionof{table}{Pass@K results of competition-level mathematical reasoning on Qwen3-4B and code generation on Seed-Coder-8B.}
    \normalsize
    \setlength{\tabcolsep}{4pt}
    \renewcommand{\arraystretch}{1.05}

    \resizebox{\linewidth}{!}{
    \begin{tabular}{@{}lccccc@{}}
    \toprule
    & \multicolumn{4}{c}{Math} & Code \\
    \cmidrule(lr){2-5} \cmidrule(lr){6-6}
    Model & AIME24 & AIME25 & AMC23 & \multirow{2}{*}{\textbf{Avg}} & LCBv6 \\
    & pass@32 & pass@32 & pass@32 & & pass@8 \\
    \midrule
    Initial & 60.00 & 53.33 & 97.50 & 70.28 & 31.97 \\
    GRPO$_{\scriptscriptstyle \mathrm{BS}=128}$ & 76.67 & 63.33 & 95.00 & 78.33 & 32.33 \\
    GRPO$_{\scriptscriptstyle \mathrm{BS}=256}$ & 73.33 & \textbf{70.00} & 95.00 & 79.44 & 32.98 \\
    GRPO$^{\scriptsize m=16}_{\scriptsize \mathrm{BS}=128}$
    & 63.33 & 60.00 & 92.5 & 71.94 & 32.61 \\
    \rowcolor{cyan!10}
    CIPO & \textbf{86.67} & \textbf{70.00} & \textbf{100.00} & \textbf{85.56} & \textbf{37.53} \\
    \bottomrule
    \end{tabular}
    }
    \label{tab:math_passk}
\end{minipage}
\hfill
\begin{minipage}[t]{0.43\linewidth}
    \vspace{0pt}
    \centering
    \includegraphics[width=\linewidth]{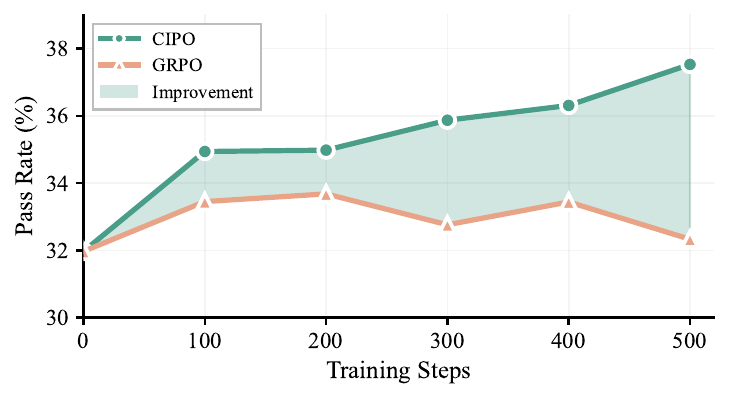}

    \captionof{figure}{Pass@8 training dynamics on LiveCodeBench v6.}
    \label{fig:code_passk}
\end{minipage}
\end{table}

\subsection{Main Results}

\begin{table}[!t]
\centering
\caption{Qwen3-4B trained with \textsc{CIPO} demonstrate improved correction and critique capabilities on CriticBench in-domain, with effective generalization to out-of-domain tasks. Comm.: Commonsense; Symb.: Symbolic; Algo.: Algorithmic.}
\label{tab:criticbench}
\normalsize
\resizebox{0.65\textwidth}{!}{

\setlength{\tabcolsep}{4pt}
\renewcommand{\arraystretch}{1.0}
\begin{tabular}{@{}l ccccc@{}}
\toprule
& {In-Domain} & \multicolumn{3}{c}{{Out-of-Domain}} & \\
\cmidrule(lr){2-2} \cmidrule(lr){3-5}
\multirow{-2}{*}{{Model}} & Math & Comm. & Symb. & Algo. & \multirow{-2}{*}{\textbf{Avg}} \\
\midrule
\rowcolor{gray!20}
\multicolumn{6}{c}{\textbf{\textit{Correction}}} \\
Qwen2.5-72B-Instruct & 73.93 & 55.27 & 92.26 & 77.30 & 74.69 \\
% Qwen3-8B & 69.63 & 49.34 & 92.26 & 78.01 & 72.31 \\
Qwen3-32B & 74.23 & 55.00 & 97.21 & 76.95 & 75.85 \\
\midrule
Initial & 67.64 & \textbf{49.07} & 86.69 & 71.63 & 68.75 \\
\quad +\textsc{GRPO} & \hspace{20pt}70.71\enspace{\color{darkgreen}\scriptsize$\uparrow$3.07} & 48.89 & 89.78 & 72.34 & 70.43 \\
\rowcolor{cyan!10}
\quad +\textsc{CIPO} & \hspace{20pt}\textbf{75.38}\enspace{\color{darkgreen}\scriptsize$\uparrow$7.74} & 48.72 & \textbf{92.72} & \textbf{74.47} & \textbf{72.82} \\
\midrule
\rowcolor{gray!20}
\multicolumn{6}{c}{\textbf{\textit{Critique}}} \\
Qwen2.5-72B-Instruct & 89.81 & 64.94 & 91.99 & 63.58 & 77.58 \\
% Qwen3-8B & 90.54 & 63.37 & 89.70 & 52.48 & 74.02 \\
Qwen3-32B & 94.10 & 67.36 & 94.41 & 56.12 & 78.00 \\
\midrule
Initial & 92.82 & 58.22 & 83.68 & 50.72 & 71.36 \\
\quad +\textsc{GRPO} & \hspace{20pt}93.87\enspace{\color{darkgreen}\scriptsize$\uparrow$1.05} & 61.55 & \textbf{90.51} & 54.79 & 75.18 \\
\rowcolor{cyan!10}
\quad +\textsc{CIPO} & \hspace{20pt}\textbf{94.00}\enspace{\color{darkgreen}\scriptsize$\uparrow$1.18} & \textbf{61.74} & 90.42 & \textbf{55.94} & \textbf{75.53} \\
\bottomrule
\end{tabular}
}
\vskip -5pt
\end{table}

\begin{table}[!t]
\centering
\normalsize
\setlength{\tabcolsep}{4pt}
\renewcommand{\arraystretch}{1.0}
\caption{Seed-Coder-8B trained with \textsc{CIPO} achieves consistent improvements in code debugging performance on DebugBench under different completeness settings.}
\label{tab:debugbench}
\resizebox{0.7\textwidth}{!}{

\begin{tabular}{@{}lc@{\hspace{8pt}}lc@{\hspace{8pt}}lc@{\hspace{8pt}}lc@{\hspace{8pt}}l@{}}
\toprule
Metric & \multicolumn{2}{c}{\hspace{-16pt}w/ Intent} & \multicolumn{2}{c}{\hspace{-16pt}w/ Full Intent} & \multicolumn{2}{c}{\hspace{-16pt}w/o Intent} & \multicolumn{2}{c}{\hspace{-16pt}\textbf{Avg}} \\
\midrule
Claude-Sonnet-4 & 66.04 && 68.17 && 62.52 && 65.58 & \\
Qwen2.5-72B-Instruct & 61.65 && 61.59 && 60.80 && 61.35 & \\
Qwen3-32B & 57.49 && 57.00 && 54.38 && 56.29 & \\
% Qwen3-8B & 50.88 && 52.36 && 44.43 && 49.22 & \\
\midrule
Initial & 61.88 && 63.99 && 56.51 && 60.79 & \\
\quad +\textsc{GRPO} & 63.25 && 66.16 && 57.97 && 62.46 & \\
\rowcolor{cyan!10}
\quad +\textsc{CIPO} & \textbf{65.33} & {\color{darkgreen}\scriptsize$\uparrow$3.45} & \textbf{68.82} & {\color{darkgreen}\scriptsize$\uparrow$4.83} & \textbf{60.82} & {\color{darkgreen}\scriptsize$\uparrow$4.31} & \textbf{64.99} & {\color{darkgreen}\scriptsize$\uparrow$4.20} \\
\bottomrule
\end{tabular}
}
\vskip -10pt
\end{table}

\textbf{\textsc{CIPO} yields significant improvements in model reasoning performance.} To validate the effectiveness of CIPO in reasoning tasks, we conduct a systematic comparison between CIPO and the strong baseline GRPO on mathematical reasoning and code generation benchmarks. As shown in Table~\ref{tab:main_results}, CIPO consistently outperforms GRPO across all reasoning tasks. Specifically, our method achieves an overall accuracy of 64.38\% on mathematical reasoning, surpassing GRPO by 4.55\%, with even larger gains on the more challenging AIME24 and AIME25 datasets, while also delivering stable improvements in code generation. Notably, under matched computational budgets, CIPO still outperforms GRPO (BS=256) by 4.72\%, which further confirms that the observed gains primarily stem from algorithmic design rather than increased computational resources.

\textbf{\textsc{CIPO} successfully expands the model's intrinsic reasoning capabilities, which vanilla GRPO struggles to achieve.} To validate the advantage of CIPO in expanding intrinsic reasoning ability, we evaluate the pass@32 metric on competition-style mathematical benchmarks and analyze the training dynamics on code generation tasks. The results demonstrate that CIPO genuinely expands the model's intrinsic capacity rather than merely reshuffling solutions via sampling. Specifically, under a fixed budget of 32 samples, CIPO outperforms vanilla GRPO by 6.12\% on mathematical tasks. Furthermore, on code generation, CIPO maintains a robust, monotonic upward trajectory throughout training, effectively preventing the performance saturation and fluctuation observed in the GRPO baseline, which further indicates that CIPO continuously explores diverse solutions to substantially enhance reasoning capacity.

\textbf{\textsc{CIPO} substantially enhances the model's correction ability.} To validate CIPO's effectiveness in error correction, we evaluate it on CriticBench and DebugBench. As shown in Table~\ref{tab:criticbench} and Table~\ref{tab:debugbench}, CIPO consistently improves error detection and rectification, significantly outperforming GRPO. Specifically, on CriticBench (Math), CIPO boosts the correction rate by 7.74\%, surpassing GRPO by 4.67\%. On DebugBench, CIPO achieves a 4.20\% gain, outperforming GRPO (+2.53\%) and even surpassing Qwen2.5-72B-Instruct while matching Claude-Sonnet-4~\citep{anthropic_claude4_news}, with consistent gains across all settings. These results demonstrate that CIPO effectively enhances the model's ability to repair errors.

\textbf{The error correction capabilities acquired through \textsc{CIPO} training demonstrate robust cross-scenario generalization to diverse reasoning tasks.} To assess the transferability of the learned capabilities, we evaluate the math-trained model on out-of-domain tasks. The results indicate that, despite being trained solely on mathematical data, CIPO generalizes effectively to unseen scenarios. Specifically, as shown in Table~\ref{tab:criticbench}, the model achieves substantial correction gains in symbolic and algorithmic reasoning. Furthermore, CIPO enhances critique performance across all out-of-domain categories, which further suggests that it fosters general critique and correction capabilities rather than task-specific overfitting, enabling effective transfer to new reasoning environments.

\begin{table}[t]
\centering
\caption{Ablation study of key components of \textsc{CIPO} on Qwen3-4B. 
Each row shows performance when one component is removed. ``w/o on-policy replay'' corresponds to the offline variant (\textsc{CIPO-Offline}) described in Section~\ref{sec:experiments}.}
\label{tab:ablation}
\setlength{\tabcolsep}{1pt}
\resizebox{0.9\columnwidth}{!}{%
\begin{tabular}{@{}lccccccc@{}}
\toprule
{Model} & AIME24 & AIME25 & AMC23 & MATH500 & Minerva & Olympiad & \textbf{Avg} \\
\midrule
Initial & 23.54 & 21.04 & 66.02 & 82.20 & 40.44 & 47.70 & 46.82 \\
{\textsc{CIPO}} & \textbf{47.50} & \textbf{44.90} & \textbf{89.61} & \textbf{92.00} & \textbf{52.57} & \textbf{59.70} & \textbf{64.38} \\
\quad {\small w/o on-policy replay}$_{\tiny \text{BS}=128}$ & 42.50 & 38.54 & 84.45 & 88.80 & 52.21 & 56.30 & 60.47 \\
\quad {\small w/o on-policy replay}$_{\tiny \text{BS}=256}$ & 46.67 & 35.73 & 84.77 & 90.00 & 50.37 & 55.11 & 60.44 \\
\quad {\small w/o Adaptive-control} & 41.98 & 34.79 & 85.00 & 91.00 & 51.47 & 56.89 & 60.19 \\
\quad {\small w/o risk-aversion} & 38.23 & 34.17 & 82.34 & 88.80 & 46.69 & 54.22 & 57.41 \\
\quad {\small w/o difficulty-aware preference} & 44.90 & 39.06 & 86.95 & 90.00 & 48.53 & 56.44 & 60.98 \\
\bottomrule
\end{tabular}%
}
\vskip -5pt
\end{table}

\subsection{Ablation Study}

We conduct ablation studies to isolate the contributions of \textsc{CIPO}'s design choices:
(i) the effect of on-policy correction versus offline replay, and
(ii) the necessity of each proposed strategy: adaptive-control, risk-aversion reward shaping and difficulty-aware preference.

% \textbf{On-policy correction is essential.}
\textbf{On-policy replay drives substantial gains over offline replay.} To disentangle the benefit of on-policy correction from merely adding offline data, we compare \textsc{CIPO} against a ``step-0 replay'' variant that utilizes offline trajectories. As shown in Table~\ref{tab:main_results}, \textsc{CIPO} significantly outperforms the variant relying solely on offline data, establishing on-policy replay as a key driver of performance improvement. Specifically, \textsc{CIPO} outperforms the offline variant by an additional 3.91\% with consistent gains across all benchmarks, confirming that the improvements stem from the on-policy mechanism's ability to dynamically correct errors and implicit credit assignment within the current policy distribution, rather than simple data augmentation.

\textbf{Adaptive-control dynamic replaying balances exploration and exploitation.} To demonstrate the importance of adaptively regulating the ratio of successful to failed trajectories in balancing exploration and exploitation, we compare the adaptive-control strategy against a fixed 1:1 replay ratio. The results show that the fixed ratio strategy lags behind the full CIPO model by 4.19\%. Specifically, the adaptive mechanism dynamically balances the proportion of successful versus failed trajectories—leveraging failed trajectories for correction learning in early stages while adjusting the proportion later to prevent policy degradation. This ensures a sustained and informative training signal, effectively exploiting the information in failed trajectories without the risk of over-correction associated with fixed ratios.

\textbf{Risk-averse reward shaping prevents over-correction and capability regression.} To validate the necessity of risk-averse reward shaping for model stability, we conduct an ablation study by removing this component. The results reveal that this removal causes the most severe performance degradation, with overall accuracy plummeting by 6.97\%. Specifically, performance drops are significant across all datasets (ranging from -3.20\% to -10.73\%), indicating that the asymmetric reward mechanism effectively balances preventing overfitting with avoiding capability regression, thereby maintaining robust reasoning capabilities while exploring correction strategies.

% \textbf{Difficulty-aware preference improves training efficiency.} Removing this component leads to consistent performance drops across all benchmarks, indicating its importance. By prioritizing samples near the model’s capability boundary, the method avoids inefficient training on overly easy or overly difficult examples.

\textbf{Difficulty-aware preference improves training efficiency.} To verify the role of difficulty-aware preference in optimizing training efficiency, we remove this component and observe model performance. The results show that removing this component leads to consistent performance declines relative to full CIPO across all benchmarks. This validates the ``zone of proximal development'' theory: by prioritizing samples at the edge of the model's capability, the strategy avoids inefficient computation on overly simple or difficult samples.

\section{Related Works}

\textbf{Process Supervision}
To address the ambiguous optimization signals caused by sparse binary rewards in RLVR, prior work introduces process supervision to provide denser feedback by evaluating intermediate steps~\citep{lightman2023let,uesato2022solving}. A common approach trains a process reward model (PRM) to assess step-level correctness for reinforcement learning or search guidance~\citep{wang2024math,luo2024improve}. Other methods employ LLM critics for corrective feedback~\citep{xie2025capo,shinn2023reflexion} or leverage environment feedback to guide credit assignment via teacher model distributions~\citep{hubotter2026reinforcement}. Recent work such as SDPO further constructs feedback-conditioned teachers from environmental feedback or self-generated trajectories.
However, these approaches typically rely on additional annotations or incur substantial computational overhead, and auxiliary models may introduce biases that harm generalization~\citep{gao2023scaling,wen2024rethinking,kim2026does}. In contrast, CIPO derives directional signals directly from failed trajectories through correction sampling, eliminating the need for external models or human annotations while preserving the simplicity of the RLVR paradigm.

\textbf{Learning from Failure} 
Failed trajectories often contain richer signals than uniform suppression can capture. In reinforcement learning, hindsight experience replay (HER)~\citep{andrychowicz2017hindsight} alleviates sparse rewards by replacing intended goals with achieved outcomes. This idea has been extended to LLMs by converting failed interactions into training data via post-hoc rewriting, often using external rewriters to correct invalid responses~\citep{zhang2025replay}.
Another line of work studies correction-oriented supervised fine-tuning (SFT), where models are trained on critique-and-revise data to improve reasoning~\citep{an2023learning,zheng2025critic,wang2025critique}. In contrast, \textsc{CIPO} removes the need for relabeling or external rewriters, and integrates correction learning directly into an on-policy RLVR framework, avoiding the distribution shift issues of offline methods.

\textbf{Self-refinement}
Recent work improves self-refinement in language models via reinforcement learning, often using multi-turn generation where models iteratively refine their own outputs~\citep{kumar2024training}. These approaches treat failed trajectories as context for subsequent attempts and primarily optimize refined outputs, sometimes at the cost of first-pass performance. Other methods incorporate external feedback to guide correction~\citep{gehring2024rlef,chen2023teaching}.
However, most operate on a single refinement trajectory per failure, without exploring nearby alternatives. In contrast, \textsc{CIPO} samples around failed trajectories and integrates the resulting signals directly into policy optimization, enabling more informative supervision without requiring multi-turn inference at test time.

\section{Conclusion}
We introduce CIPO, which transforms on-policy failed trajectories into exploitable supervisory signals for RLVR training. The key insight is that correction naturally differentiates failure modes: near-miss attempts are more likely to yield correct solutions during refinement, enabling richer learning signals from failures. Experiments across 11 benchmarks demonstrate that CIPO achieves significant improvements on both reasoning and correction tasks, with pass@K gains indicating genuine expansion of intrinsic reasoning capabilities.

\section*{Impact Statement}

This paper presents work whose goal is to advance the field of  Machine Learning. There are many potential societal consequences  of our work, none of which we feel must be specifically highlighted here.

%Bibliography
\bibliographystyle{unsrt}  
\bibliography{references}

\clearpage
\appendix

\section{Details about Method}
\label{appendix:Details-about-method}

\subsection{Algorithm of CIPO}
Algorithm~\ref{alg:cipo} outlines the core workflow of CIPO. In practice, to improve training efficiency, correction rollouts are based on the previous step, enabling parallel sampling instead of sequentially waiting for the base rollout to complete and be rewarded.
\begin{algorithm}[H]
\caption{Correction-Oriented Policy Optimization}
\label{alg:cipo}
\begin{algorithmic}[1]
\Require Initial policy $\pi_\theta$, prompt set $\mathcal{D}$, reward function $R(\cdot)$
\State \textbf{Hyperparams:} group size $G$, replay fraction $\gamma$, difficulty range $[\delta_{\text{low}},\delta_{\text{high}}]$, risk penalty $\lambda_{\text{risk}}$, target reward $R^*$
\State Initialize $\rho \leftarrow \rho_0$, $c \leftarrow 0$, $R_{\text{prev}} \leftarrow \textbf{None}$
\For{training step $t=1,2,\ldots$}
    \State \textcolor{OliveGreen}{\textit{// Base rollouts}}
    \State Sample prompts $\{x_j\}_{j=1}^{B}$ from $\mathcal{D}$; for each $x$, sample $G$ responses and compute rewards
    \State Collect base trajectories $\mathcal{B}_{\text{base}} = \{(x, y_i, r_i)\}$
    \State \textcolor{OliveGreen}{\textit{// Correction rollouts}}
    \State $\mathcal{S}_{\text{replay}} \leftarrow \textsc{RolloutReplay}(\mathcal{B}_{\text{base}}, [\delta_{\text{low}},\delta_{\text{high}}], \lfloor\gamma B\rfloor, \rho)$ as in Algorithm~\ref{alg:replay}
    \For{each $(x, y_c, r_c) \in \mathcal{S}_{\text{replay}}$}
        \State Construct $x' \leftarrow \textsc{Augment}(x, y_c)$; sample $G$ responses $\{y'_i\}_{i=1}^G$
        \State Compute shaped rewards $\tilde{r}_i \leftarrow R_{\text{risk}}(x, y_c, y'_i)$ via Eq.~\ref{eq:risk_shaping}
    \EndFor
    \State Collect correction trajectories $\mathcal{B}_{\text{cor}} = \{(x', y'_i, \tilde{r}_i)\}$
    \State \textcolor{OliveGreen}{\textit{// Policy update}}
    \State Update $\pi_\theta$ on $\mathcal{B}_{\text{base}} \cup \mathcal{B}_{\text{cor}}$
    \State \textcolor{OliveGreen}{\textit{// Adaptive ratio update}}
    \State $\mathcal{S}_{+} \leftarrow \{(x', y'_i, \tilde{r}_i) \in \mathcal{B}_{\text{cor}} : r_c = 1\}$
    \State \textcolor{OliveGreen}{\textit{// corrections conditioned on successful trajectories}}
    \State $R_t \leftarrow \frac{1}{|\mathcal{S}_{+}|}\sum_{(x', y', \tilde{r}) \in \mathcal{S}_{+}} \tilde{r}$
    \State $\rho, c \leftarrow \textsc{UpdateRatio}(\rho, R_t, R_{\text{prev}}, c; R^*)$ as in Algorithm~\ref{alg:update_ratio}
    \State $R_{\text{prev}} \leftarrow R_t$
\EndFor
\State \textbf{Return:} $\pi_\theta$
\end{algorithmic}
\end{algorithm}

\subsection{Algorithm of RolloutReplay and UpdateRatio}

This section presents detailed descriptions of the two sub-algorithms employed in Algorithm~\ref{alg:cipo}.

\textbf{RolloutReplay} (Algorithm~\ref{alg:replay}) selects trajectories for correction rollouts. It first prioritizes prompts within the medium-difficulty range $[\delta_{\text{low}}, \delta_{\text{high}}]$ based on their empirical pass rates, then falls back to remaining prompts if needed. The selected trajectories are split into successful and failed groups, with the ratio $\rho$ controlling their mixture. The detailed hyperparameter settings are provided in Appendix~\ref{appendix:hyperparams}.

\begin{algorithm}[H]
\caption{\textsc{RolloutReplay}($\mathcal{B}$, $[\delta_{\text{low}},\delta_{\text{high}}]$, $N$, $\rho$)}
\label{alg:replay}
\begin{algorithmic}[1]
\State \textbf{Input:} base rollouts $\mathcal{B}=\{(x,y,r)\}$, difficulty range $[\delta_{\text{low}},\delta_{\text{high}}]$, target size $N$, positive ratio $\rho$
\State Compute per-prompt pass rate $\hat{P}(x) = \frac{1}{|\{(x,y,r)\in\mathcal{B}\}|}\sum_{(x,y,r)\in\mathcal{B}} r$
\State $\mathcal{B}_{\text{med}} \leftarrow \{(x,y,r)\in\mathcal{B}: \delta_{\text{low}} \le \hat{P}(x) \le \delta_{\text{high}}\}$ \textcolor{OliveGreen}{\textit{// medium-difficulty}}
\State $\mathcal{B}' \leftarrow \textsc{Shuffle}(\mathcal{B}_{\text{med}}) \oplus \textsc{Shuffle}(\mathcal{B} \setminus \mathcal{B}_{\text{med}})$ \textcolor{OliveGreen}{\textit{// prioritize medium, fallback to rest}}
\State Split $\mathcal{B}'$ into $\mathcal{B}_+ = \{(x,y,r) : r=1\}$ and $\mathcal{B}_- = \{(x,y,r) : r=0\}$
\State $N_+ \leftarrow \min(\lfloor \rho N \rfloor, |\mathcal{B}_+|)$, \quad $N_- \leftarrow \min(N - N_+, |\mathcal{B}_-|)$
\State Backfill from $\mathcal{B}_+$ if $N_+ + N_- < N$
\State \textbf{Return:} $\mathcal{B}_+[1\!:\!N_+] \cup \mathcal{B}_-[1\!:\!N_-]$
\end{algorithmic}
\end{algorithm}

\textbf{UpdateRatio.}
\label{appendix:updateratio}
As shown in Algorithm~\ref{alg:update_ratio}, we adapt the replay ratio $\rho_t \in [\rho_{\min}, \rho_{\max}]$ according to the model's retention performance on replayed successful samples. Let $R_t$ denote the average shaped reward on corrections replayed from successful trajectories at iteration $t$, and let $R^*$ denote the target retention level. The replay ratio is updated using three signals: the current performance gap to the target, the recent performance deterioration relative to the previous step, and a capped consecutive-underperformance term:
\begin{equation}
\rho_{t+1} =
\operatorname{clip}\!\left[
\rho_t \left(
1
+ w_1 (R^* - R_t)
+ w_2 \max(0, R_{t-1} - R_t)
+ w_3 \min(c_t, 3)
\right),
\, \rho_{\min}, \rho_{\max}
\right].
\end{equation}
Here, $c_t$ denotes the consecutive-underperformance counter, which is incremented when $R_t < R^*$ and reset otherwise. The first term increases the replay fraction of successful trajectories when the current retention performance falls below the target. The second term further increases this fraction when retention performance deteriorates relative to the previous iteration. The third term makes the update more conservative under persistent underperformance, while capping its contribution to avoid dominating the overall update. Detailed hyperparameter settings are provided in Appendix~\ref{appendix:hyperparams}.

\begin{algorithm}[H]
\caption{\textsc{UpdateRatio}($\rho_t$, $R_t$, $R_{t-1}$, $c_{t-1}$; $R^*$)}
\label{alg:update_ratio}
\begin{algorithmic}[1]
\State \textbf{Input:} current replay ratio $\rho_t$, current average shaped reward $R_t$, previous average shaped reward $R_{t-1}$, previous underperformance counter $c_{t-1}$, target reward $R^*$
\State \textbf{Hyperparameters:} $w_1, w_2, w_3$, $\rho_{\min}, \rho_{\max}$
\If{$R_t < R^*$}
    \State $c_t \gets c_{t-1} + 1$
\Else
    \State $c_t \gets 0$
\EndIf
\State $f_1 \gets R^* - R_t$
\If{$R_{t-1}$ is available}
    \State $f_2 \gets \max(0, R_{t-1} - R_t)$
\Else
    \State $f_2 \gets 0$
\EndIf
\State $f_3 \gets \min(c_t, 3)$
\State $\rho_{t+1} \gets \textsc{Clip}\!\left(\rho_t \cdot (1 + w_1 f_1 + w_2 f_2 + w_3 f_3),\, \rho_{\min},\, \rho_{\max}\right)$
\State \Return $\rho_{t+1}, c_t$
\end{algorithmic}
\end{algorithm}

\subsection{Correction Prompt Construction.}
\label{sec:Correction Prompt Construction}
After selecting a subset of rollouts $\mathcal{R}$ (containing both correct and incorrect candidates), we construct \textbf{correction prompts} that condition on the original problem and the model's previous attempt. Specifically, for a selected rollout $(x, y_{\text{cand}}, r)$, we create:

\begin{tcolorbox}[
    colback=gray!10,      
    colframe=gray!50,     
    boxrule=0.5pt,        
    arc=2pt,             
    left=5pt, right=5pt,  
    top=5pt, bottom=5pt
]
\small\ttfamily
\{Original Prompt x\} \\[0.5em]
Below is a candidate solution from a large language model (\textbf{correctness unknown}): \\[0.5em]
<candidate\_solution> \\
\{y\_cand\} \\
</candidate\_solution> \\[0.5em]
Please refer to this solution and provide your solution.
\end{tcolorbox}

Crucially, we inform the model that the candidate's \textbf{correctness is unknown}, avoiding explicit labeling that would leak reward information and undermine the RL objective.

This construction transforms a potentially failed response into a structured learning signal: the model learns to \textit{recognize and fix its own mistakes} (when $y_{\text{cand}}$ is incorrect) or \textit{reinforce correct reasoning patterns} (when $y_{\text{cand}}$ is correct), rather than simply avoiding failures through likelihood suppression.

\section{Hyperparameters}
\label{appendix:hyperparams}
Table~\ref{tab:hyperparams} details the full hyperparameters configuration for CIPO.

\begin{table}[ht]
\renewcommand{\arraystretch}{1.1}
\begin{center}
\begin{tabular}{lll}
\toprule
Module & Hyperparameter & Value \\
\midrule

Adaptive Replay Ratio
& $\rho_0$ & 0.3 \\
& $\rho_{\min}, \rho_{\max}$ & $0.2, 0.8$ \\
& $w_1, w_2, w_3$ & $0.8, 0.3, 0.05$ \\
& $R^*$ & 0.80 \\

Risk-Averse Reward Shaping
& $\lambda_{\text{risk}}$ & 1 \\

Difficulty-aware Trajectory Preference
& $\delta_{\text{low}}, \delta_{\text{high}}$ & $\frac{3}{8}, \frac{6}{8}$ \\

Replay Sampling
& $n$ & 8 \\

Joint Optimization
& $\lambda$ & 1 \\

\bottomrule
\end{tabular}    
\end{center}
\caption{Hyperparameter configuration of CIPO.}
\label{tab:hyperparams}
\end{table}

\end{document}